\documentclass{article}

\usepackage[nonatbib, preprint]{neurips_2024}
\usepackage{multirow}

\usepackage[markup=underlined]{changes}

\usepackage[utf8]{inputenc}   
\usepackage[T1]{fontenc}    	
\usepackage{url}              		 
\usepackage{booktabs}       	
\usepackage{amsfonts}       	
\usepackage{nicefrac}       	  
\usepackage{microtype}      	

\usepackage{graphicx}
\usepackage{color}
\usepackage{rotating}
\usepackage{tabularx}
\usepackage{pdflscape}
\usepackage{amsmath,amsthm,amssymb}

\usepackage{algorithm}
\usepackage[noend]{algorithmic}

\usepackage{bbm,dsfont}
\usepackage{subfig}
\usepackage{caption}
\usepackage{wrapfig}
\usepackage{cancel}
\usepackage{enumerate,cases}
\usepackage{thmtools,thm-restate}
\usepackage{mathtools}

\usepackage[none]{hyphenat}
\usepackage{multirow}
\usepackage{makecell}
\usepackage{setspace}
\usepackage{pifont}
\usepackage{array}
\usepackage{enumitem}
\usepackage{lineno}     

\allowdisplaybreaks

\usepackage{hyperref}		 
\usepackage[capitalize]{cleveref}

\usepackage{todonotes}

\let\svthefootnote\thefootnote
\newcommand\blankfootnote[1]{%
  \let\thefootnote\relax\footnotetext{#1}%
  \let\thefootnote\svthefootnote%
}

\newcommand{\Prob}[1]{\bP\left\{#1\right\}}

\renewcommand{\phi}{\varphi}
\renewcommand{\epsilon}{\varepsilon}

\newcommand{\norm}[1]{\left\|#1\right\|}

\newcommand{\el}{\end{flushleft}}
\newcommand{\bl}{\begin{flushleft}}

\newcommand{\argmax}{\arg\!\max}

\newcommand{\bP}{\mathbb{P}}

\theoremstyle{plain}

\usepackage{selectp}

\usepackage{tcolorbox}
\usepackage{colortbl}
\usepackage{wrapfig}
\newtcolorbox[auto counter, number freestyle={\noexpand\arabic{\tcbcounter}}]{mycolorbox}[3][]{%
    fonttitle=\bfseries,
    title=Example~#2~\thetcbcounter: #3,
    #1
}

\newcommand{\alg}{\texttt{APOHF}}

\title{
    Prompt Optimization with Human Feedback
}

\author{
Xiaoqiang Lin$^{1}$, Zhongxiang Dai$^{2}$\thanks{Corresponding author.}, Arun Verma$^{1}$, See-Kiong Ng$^{1,3}$, \\
\textbf{Patrick Jaillet}$^{2}$, \textbf{Bryan Kian Hsiang Low}$^{1}$\\
$^{1}$Department of Computer Science, National University of Singapore\\
$^{2}$LIDS and EECS, Massachusetts Institute of Technology\\
$^{3}$Institute of Data Science, National University of Singapore\\
\texttt{xiaoqiang.lin@u.nus.edu, daizx@mit.edu, arun@comp.nus.edu.sg},\\
\texttt{seekiong@nus.edu.sg, jaillet@mit.edu, lowkh@comp.nus.edu.sg}
}

\begin{document}    
    \maketitle
        
    \begin{abstract}
        Large language models (LLMs) have demonstrated remarkable performances in various tasks. However, the performance of LLMs heavily depends on the input prompt, which has given rise to a number of recent works on \emph{prompt optimization}. However, previous works often require the availability of a numeric score to assess the quality of every prompt. Unfortunately, when a human user interacts with a black-box LLM, attaining such a score is often infeasible and unreliable. Instead, it is usually significantly easier and more reliable to obtain \emph{preference feedback} from a human user, i.e., showing the user the responses generated from a pair of prompts and asking the user which one is preferred. Therefore, in this paper, we study the problem of \emph{prompt optimization with human feedback} (POHF), in which we aim to optimize the prompt for a black-box LLM using only human preference feedback. Drawing inspiration from dueling bandits, we design a theoretically principled strategy to select a pair of prompts to query for preference feedback in every iteration, and hence introduce our algorithm named \emph{automated POHF} (\alg). We apply our \alg~algorithm to various tasks, including optimizing user instructions, prompt optimization for text-to-image generative models, and response optimization with human feedback (i.e., further refining the response using a variant of our \alg). The results demonstrate that our \alg~can efficiently find a good prompt using a small number of preference feedback instances.
        Our code can be found at \url{https://github.com/xqlin98/APOHF}.
    \end{abstract}

    \section{Introduction}
    \label{sec:introduction}

Large language models (LLMs) have shown impressive performances in a variety of tasks \cite{ArXiv23_google2023palm,ArXiv23_openai2023gpt}.
However, the performances of LLMs are significantly dependent on the \emph{prompt} given to them \cite{ICLR23_zhou2022large}.
Unfortunately, finding the best prompt for an LLM to perform a task is often challenging, especially considering that the most powerful LLMs nowadays are often \emph{black-box} models to which only API access is available \cite{ArXiv23_openai2023gpt}.
This challenge has given rise to a number of recent works on \emph{prompt optimization} for black-box LLMs, which aim to efficiently find the best prompt for a black-box LLM 
\cite{ArXiv23_chen2023instructzero, ArXiv23_lin2023use,ICLR23_zhou2022large}.
These works have shown that prompt optimization
can dramatically improve the performances of black-box LLMs in various tasks.
However, these works often impose a potentially unrealistic requirement on the tasks: \emph{They usually require access to a numeric score to evaluate the performance of every prompt}.
This significantly limits their practicality in real-world use cases.

Specifically, some works on prompt optimization have assumed the availability of a validation set, which can be used to evaluate (the response generated from) a candidate prompt \cite{ArXiv23_chen2023instructzero,hu2024localized,ArXiv23_lin2023use}.
Meanwhile, other works have used a separate LLM (often referred to as the scorer LLM) to provide a score indicating the efficacy of (the response produced by) a prompt \cite{yang2023large,ICLR23_zhou2022large}.
However, \emph{when a human user directly interacts with a black-box LLM to perform a task} (i.e., the most common use cases of LLMs nowadays), these methods to obtain a score are often unrealistic.
This is because in such use cases, a validation set is usually unavailable and the scorer LLM is unlikely to provide an accurate assessment of a prompt for the task the user has in mind.
Therefore, these previous prompt optimization methods are inapplicable for such use cases.
In addition, directly asking a user for a numeric score to assess (the response generated by) a candidate prompt is usually infeasible and unreliable \cite{yue2012k}.
Instead, a human user is often significantly more willing to and reliable at providing \emph{preference feedback}, i.e., examining the responses generated by a pair of prompts and indicating which one is preferred \cite{yue2012k}.
This naturally begs the question: \textbf{Can we achieve prompt optimization using only human preference feedback?}
In this work, we tackle this important problem, which we refer to as \emph{prompt optimization with human feedback} (POHF).

The significance of POHF can also be highlighted by drawing an analogy to \emph{reinforcement learning with human feedback} (RLHF) \cite{ziegler2019fine}.
RLHF, as well as its variants such as direct preference optimization \cite{rafailov2024direct}, uses a dataset of human preference feedback 
to fine-tune the parameters of an LLM in order to align the LLM with human values \cite{rafailov2024direct}.
The tremendous success of RLHF is evidence of the advantage of using human preference feedback to adapt LLMs.
While RLHF has relied on fine-tuning the model parameters to adapt the response of an LLM (to align with human values), 
our POHF aims to \emph{use prompt optimization to adapt the response of an LLM to perform a task for a human}.
Interestingly, our algorithm for POHF can be extended to further refine the response of an LLM through \emph{response optimization with human feedback} (Sec.~\ref{subsec:exp:response:opt}).
Specifically, for every received prompt,
we can use the LLM to generate a large pool of responses and then strategically select a pair of responses from the pool to query for user preference feedback \cite{arXiv23_dwaracherla2024efficient}.
Our goal here is to find the best response for every given prompt while using only human preference feedback.
This can be useful in applications where we do not have the flexibility to choose the prompt,
but can sample a large number of responses from the LLM.
For example, it may be adopted by an LLM provider to further refine its response to user prompts while only collecting user preference feedback.

\begin{figure}
    \centering
  \begin{minipage}[b]{0.55\textwidth}
    \centering
    \includegraphics[width=0.95\linewidth]{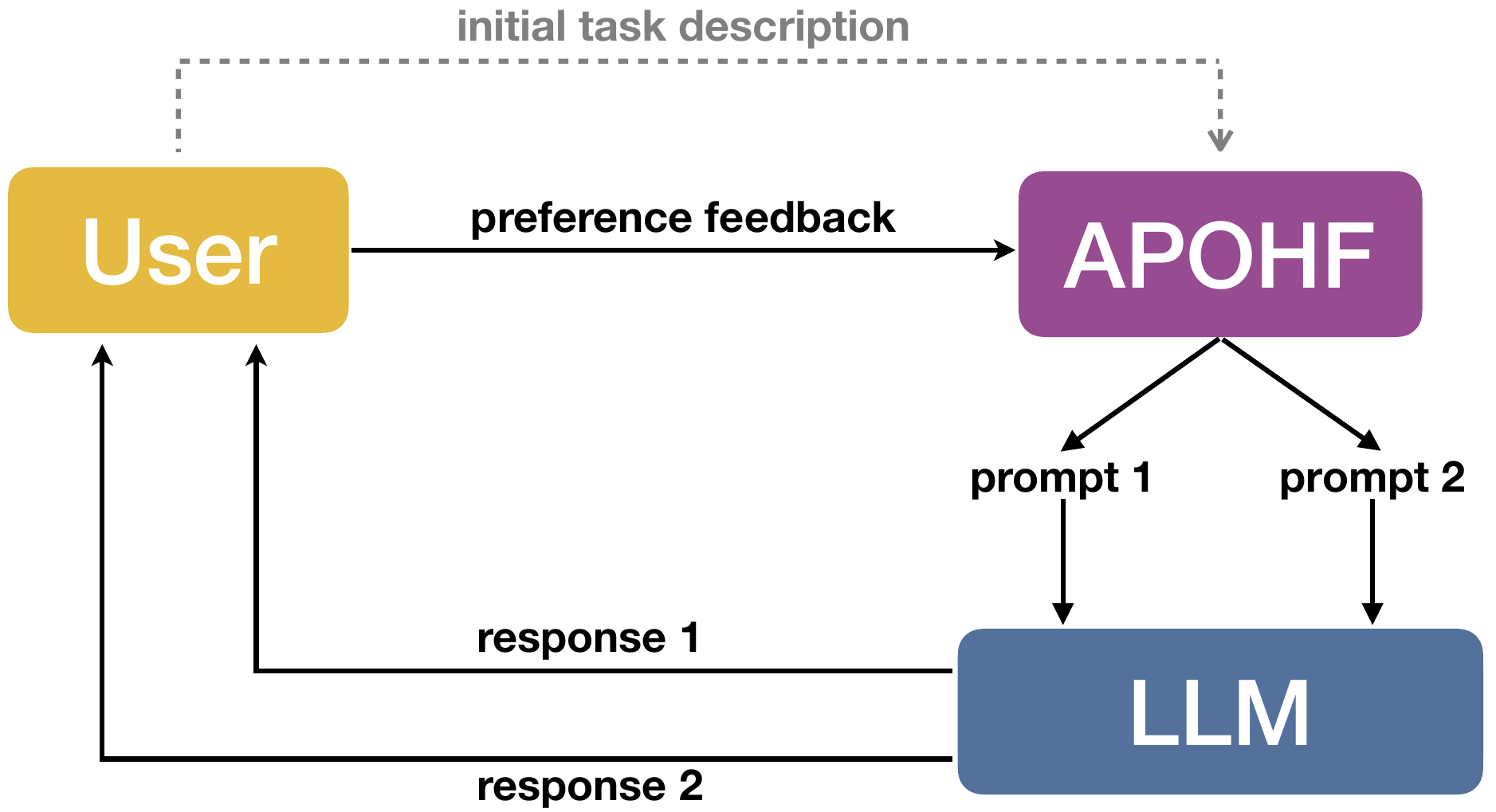}
    \caption{Illustration of our automated prompt optimization with human feedback (\alg).}
    \label{fig:problem:setting}
  \vspace{-5mm}
  \end{minipage}
  \hfill
  \begin{minipage}[b]{0.4\textwidth}
    \centering
    \includegraphics[width=0.75\linewidth]{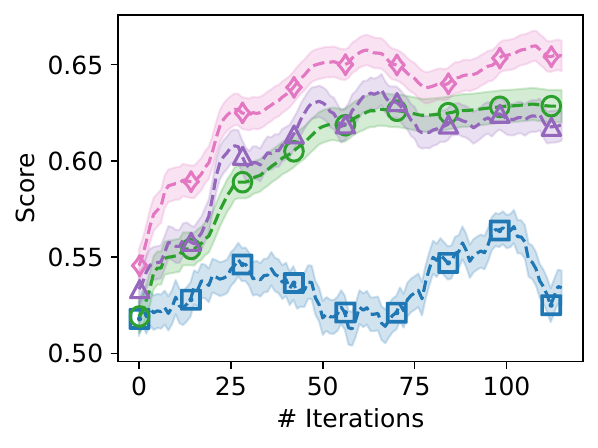}
    \\
    \includegraphics[width=0.93\linewidth]{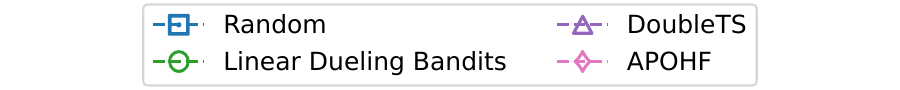}
    \caption{Latent scores of different methods in user instruction optimization, averaged over 30 tasks (Sec.~\ref{subsec:exp:instruction}).}
    \label{fig:benchmark-db}
  \vspace{-5mm}
  \end{minipage}
\end{figure}

Similar to RLHF, in our POHF, it is of paramount importance to find a good prompt \emph{using a small number of human feedback instances}.
This is because collecting human feedback can usually be expensive and time-consuming.
To achieve this, inspired by \cite{ArXiv23_lin2023use}, we adopt the embedding from a pre-trained language model as the continuous representation of the prompts, and train a neural network (NN), which takes the embedding as input, to predict the performance (i.e., the latent score, see Sec.~\ref{subsec:method:train:nn}) of different prompts.
Based on the trained NN, we draw inspiration from \emph{dueling bandits} \cite{bengs2022stochastic,saha2021optimal} and design a theoretically principled strategy to select the pair of prompts (to query for human feedback) in every iteration.
Specifically, we choose the first prompt following a greedy strategy, i.e., by selecting the prompt that is predicted to have the best performance by the trained NN. Next, we select the second prompt based on the principle of upper confidence bound, which allows us to simultaneously \emph{exploit} the performance prediction from the NN and \emph{explore} those prompts whose performance prediction has large uncertainty.
As a result of the accurate performance prediction of the NN (thanks to the expressive power of the pre-trained embedding and the NN) and our principled prompt selection strategy, our algorithm, named \emph{Automated POHF} (\alg), is able to find a good prompt using only a small number of human preference feedback instances.

Within our problem setting (illustrated in Fig.~\ref{fig:problem:setting}), our \alg~algorithm acts as an interface between the user and the LLM.
To adopt our \alg~in practice, the user only needs to provide (1) an initial task description (e.g., a few input-output exemplars or an initial prompt) and subsequently (2) a series of preference feedback between pairs of responses (more details in Sec.~\ref{subsec:method:user:pref:feedback}).
We adopt a number of tasks to validate the performance of our \alg, including optimizating user instructions (Sec.~\ref{subsec:exp:instruction}), prompt optimization for text-to-image generative models (Sec.~\ref{subsec:exp:image}), and response optimization with human feedback (Sec.~\ref{subsec:exp:response:opt}).
In these tasks, our \alg~consistently achieves better performances than baseline methods, demonstrating its immense potential in real-world applications.

    \section{Problem Setting}
    \label{sec:problem}

In POHF, we aim to find a prompt $x\in\mathcal{X}$ that maximizes an unknown function $u$, which we refer to as the 
latent score/utility function.
That is, we aim to solve the following optimization problem: $x^\star = \argmax_{x \in \mathcal{X}} u(x)$
while only observing human \emph{preference feedback}. 
In every iteration $t$, we select a pair of prompts $x_{t,1}$ and $x_{t,2}$ to obtain their corresponding LLM-generated responses and show them to the user.
Then, we collect a binary observation $y_t = \mathbbm{1}(x_{t,1} \succ x_{t,2})$, which is equal to $1$ if the human user prefers the response from $x_{t,1}$ over that from $x_{t,2}$ and $0$ otherwise.
To model the preference feedback, we adopt the commonly used Bradley-Terry-Luce (BTL) model \cite{AS04_hunter2004mm}.
That is, for any pair of prompts $x_1$ and $x_2$, the probability that $x_1$ is preferred over $x_2$ is given by $\mathbb{P}(x_{1} \succ x_{2}) = \sigma (u(x_{1}) - u(x_{2}))$, in which $\sigma(\cdot)$ denotes the logistic function: $\sigma(x) ={1}/{(1+\mathrm{e}^{-x})}$.
The binary observation $y_t$ is then sampled from a Bernoulli distribution with probability $\mathbb{P}(x_{1} \succ x_{2})$.
The stochastic nature of $y_t$ allows us to naturally account for the noise in human preferences between a pair of prompts.
The noise may arise due to different sources of randomness, such as the randomness in the LLM-generated response for a given prompt, the variability in human decisions, among others.

Following 
recent works on query-efficient prompt optimization \cite{ArXiv23_chen2023instructzero,ArXiv23_lin2023use}, we convert POHF into a continuous optimization problem.
Specifically, for every prompt $x\in\mathcal{X}$ in the domain,
we extract the embedding from a pre-trained language model as its continuous representation. 
Of note, the previous works of \cite{ArXiv23_chen2023instructzero} and \cite{ArXiv23_lin2023use} adopted a separate white-box LLM so that the soft prompt (i.e., a part of the input to the white-box LLM to generate the prompt) can be used as the continuous representation of the prompt.
Therefore, compared to \cite{ArXiv23_chen2023instructzero} and \cite{ArXiv23_lin2023use}, our method of adopting the embedding from a pre-trained model removes the need for the white-box LLM, and hence significantly reduces the complexity and computational cost.
To simplify notations, hereafter, we use $x$ to denote the continuous embedding of a prompt in the domain.
Before the beginning of our algorithm, we use the initial task description from the user (Fig.~\ref{fig:problem:setting}) to generate the discrete domain of prompts $\mathcal{X}$, which we discuss in more detail in Sec.~\ref{subsec:method:user:pref:feedback}.

\vspace{-2mm}

    \section{Automated Prompt Optimization with Human Feedback (\alg)}
    \label{sec:pohf}

\textbf{Overview of \alg~(illustrated in Fig.~\ref{fig:problem:setting}).}
In every iteration $t$ of our \alg~algorithm (Algo.~\ref{algo:apohf}), we firstly use the current history of preference observations $\mathcal{D}_{t-1}=\{(x_{s,1}, x_{s,2}, y_s)\}_{s=1,\ldots,t-1}$ to train a neural network (NN) for score prediction (Sec.~\ref{subsec:method:train:nn}). 
Next, we leverage the trained NN to select the next pair of prompts $x_{t,1}$ and $x_{t,2}$ to query (Sec.~\ref{subsec:method:selecting:pair:of:prompts}).
Then, the pair of prompts $x_{t,1}$ and $x_{t,2}$ are used to generate their respective responses, which are shown to the human user who gives preference feedback $y_t = \mathbbm{1}(x_{t,1} \succ x_{t,2})$ (Sec.~\ref{subsec:method:user:pref:feedback}).
The newly collected observation $(x_{t,1}, x_{t,2}, y_t)$ is then added to the history, which is subsequently used to train the NN for the next iteration $t+1$.
\vspace{-1mm}

\begin{algorithm}
	\begin{algorithmic}[1]
		\FOR{$t= 1, \ldots, T$}
            \STATE Train NN using history $\mathcal{D}_{t-1}=\{(x_{s,1}, x_{s,2}, y_s)\}_{s=1,\ldots,t-1}$ by minimizing loss function \eqref{eq:customized:loss:function}
            \STATE Choose the first prompt $x_{t,1}$ by maximizing the NN prediction 
            \STATE Choose the second prompt $x_{t,2}$ by maximizing the upper confidence bound in \cref{eq:choose:arm:2}
		\STATE Obtain the responses from $x_{t,1}$ and $x_{t,2}$, and observe user preference: $y_t = \mathbbm{1}(x_{t,1}\succ x_{t,2})$
		\ENDFOR
        \STATE Train NN using entire history, report $x^*_T = {\arg\max}_{x\in\{x_{s,1},x_{s,2}\}_{s=1,\ldots,T}} h(x;\theta_T)$ as best prompt
	\end{algorithmic}
\caption{Automated Prompt Optimization with Human Feedback (\alg)}
\label{algo:apohf}
\end{algorithm}
\vspace{-1mm}

\vspace{-1mm}
\subsection{Training the Neural Network for Latent Score Prediction}
\vspace{-1mm}
\label{subsec:method:train:nn}
In our \alg, we adopt an NN (more specifically, a multi-layer perceptron, or MLP) with parameters $\theta$, denoted as $h(x;\theta)$.
The NN takes as input the pre-trained embedding $x$ of a prompt and predicts its latent score $u(x)$.
Therefore, for a pair of prompts $x_1$ and $x_2$, we use $\sigma \left( h(x_1;\theta) - h(x_2;\theta) \right)$ to model the probability that $x_1$ is preferred over $x_2$: $\mathbb{P}(x_{1} \succ x_{2}) = \sigma (u(x_{1}) - u(x_{2}))$.

In iteration $t$, given the current history of preference observations $\mathcal{D}_{t-1}=\{(x_{s,1}, x_{s,2}, y_s)\}_{s=1,\ldots,t-1}$, we train the NN using gradient descent to minimize the following loss function:
\begin{equation}
\resizebox{0.99\linewidth}{!}{
$
\mathbf{l}_t(\theta) = -\left(\sum^{t-1}_{s=1} \Big[ y_s\log \sigma \big( h(x_{s,1};\theta) - h(x_{s,2};\theta) \big) +  (1 - y_s) \log \sigma \big( h(x_{s,2};\theta) - h(x_{s,1};\theta) \big) \Big] \right) + \lambda \norm{\theta}^2_{2}.
$
}
\label{eq:customized:loss:function}
\end{equation}
Recall that $y_s = \mathbbm{1}(x_{s,1} \succ x_{s,2})$.~Intuitively, minimizing this loss function \eqref{eq:customized:loss:function} corresponds to obtaining the maximum log-likelihood estimate of the MLP parameters $\theta$ (with L2 regularization) using the preference dataset $\mathcal{D}_{t-1}$.
The strong expressive power of the pre-trained embedding and the NN helps us accurately estimate the latent score function $u$, which is crucial for the strong performance of our \alg~algorithm.
After the NN is trained, the resulting NN with parameters $\theta_t={\arg\min}_{\theta} \mathbf{l}_t(\theta)$ is used to select the pair of prompts to query in iteration $t$ (Sec.~\ref{subsec:method:selecting:pair:of:prompts}).

\vspace{-1mm}
\subsection{Selecting the Next Pair of Prompts}
\vspace{-1mm}
\label{subsec:method:selecting:pair:of:prompts}
The prompt selection strategy of our \alg~is designed by drawing inspirations from the theoretically principled \emph{linear dueling bandits} \cite{bengs2022stochastic,saha2021optimal}.
However, note that instead of using a linear model to learn the score function \cite{bengs2022stochastic,saha2021optimal}, we adopt an NN (Sec.~\ref{subsec:method:train:nn}) to make our \alg~not only theoretically grounded but also practically effective.
As we verify in Sec.~\ref{sec:experiments}, our \alg~substantially outperforms linear dueling bandits in all our experiments.
We also provide some high-level theoretical justifications for our prompt selection strategy in App.~\ref{app:sec:theoretical:justification}.

We choose \textbf{the first prompt} greedily, i.e., by selecting the one predicted to have the largest latent score using the trained NN (Sec.~\ref{subsec:method:train:nn}): $x_{t,1} = \arg\max_{x\in\mathcal{X}} h(x;\theta_t)$.
Next, after the first prompt $x_{t,1}$ is selected, we choose \textbf{the second prompt} $x_{t,2}$ by maximizing an upper confidence bound:
\begin{equation}
\begin{split}
x_{t,2} = \arg\max_{x\in\mathcal{X}}h(x;\theta_t) + \nu \norm{\nabla h(x;\theta_t) - \nabla h(x_{t,1};\theta_t) }_{V_{t-1}^{-1}},
\end{split}
\label{eq:choose:arm:2}
\end{equation}
in which 
$V_t = \sum_{s=1}^t \phi'_s {\phi'}_s^\top + \lambda \mathbf{I}$, and $\phi'_s = \nabla h(x_{s,1};\theta_s) - \nabla h(x_{s,2};\theta_s)$.
Our strategy to select the second prompt \eqref{eq:choose:arm:2} is able to balance the exploration-exploitation trade-off.
Specifically, the first term $h(x;\theta_t)$ allows us to \textbf{exploit} the predicted score of the trained NN.
Meanwhile, the second term in \eqref{eq:choose:arm:2} characterizes our \emph{uncertainty} about the score of $x$ given (a) the 
prompts selected in the previous iterations $\mathbf{X}_{t-1}=\{(x_{s,1},x_{s,2})\}_{s=1,\ldots,t-1}$ and (b) the first selected prompt $x_{t,1}$.
Intuitively, a larger value of the second term (i.e., a larger uncertainty) suggests that $x$ is more different from the previously queried prompts $\mathbf{X}_{t-1}$ and the first selected prompt $x_{t,1}$.
Therefore, maximizing the second term in \eqref{eq:choose:arm:2} helps us \textbf{explore} the domain of prompts by promoting the selection of a prompt that is different from the previously selected prompts (including those in $\mathbf{X}_{t-1}$ and $x_{t,1}$).
Here, $\nu$ is a parameter that controls the trade-off between exploration and exploitation.

In addition to being theoretically principled, another advantage of our prompt selection strategy is that it provides us with a natural method to choose the prompt to report as the best prompt.
In 
POHF, we only have access to binary preference feedback between pairs of prompts and cannot observe numeric scores indicating the efficacy of different prompts.
Therefore, it is non-trivial to choose which prompt to recommend as the best prompt.
Interestingly, our strategy to select the first prompt provides a natural and principled way to choose the prompt to recommend.
Specifically, after any iteration, we train the NN using the current history of preference observations, and choose the prompt (among all previously selected prompts) which maximizes the predicted score of the trained NN to report as the best prompt (line 6 of Algo.~\ref{algo:apohf}).
This is in fact analogous to a common practice in Bayesian optimization, i.e., choosing the input (among all previously queried inputs) that maximizes the predicted function value (i.e., the Gaussian process posterior mean) to report as the best input~\cite{nguyen2021value}.

\vspace{-1mm}
\subsection{Collecting User Preference Feedback}
\label{subsec:method:user:pref:feedback}
\vspace{-1mm}
After the pair of prompts $x_{t,1}$ and $x_{t,2}$ are selected, we then separately pass them to the target black-box LLM to produce their corresponding responses.
Next, these two responses are shown to the user, who then gives preference feedback $y_t = \mathbbm{1}(x_{t,1},x_{t,2})$  
indicating which one of the two responses (generated from $x_{t,1}$ and $x_{t,2}$) is preferred.
Then, the newly collected observation $(x_{t,1}, x_{t,2}, y_t)$ is added to the history of preference observations to yield $\mathcal{D}_{t}=\{(x_{s,1}, x_{s,2}, y_s)\}_{s=1,\ldots,t}$, after which we use the updated history $\mathcal{D}_{t}$ to train our NN (Sec.~\ref{subsec:method:train:nn}) and proceed to the next iteration $t+1$.

In addition to the 
above-mentioned 
preference feedback,
at the beginning of our \alg, the user needs to provide some initial task description (Fig.~\ref{fig:problem:setting}), which our \alg~algorithm uses to generate the domain of prompts (Sec.~\ref{sec:problem}).
The initial task description may be in the form of some input-output exemplars for the task (we follow this in our experiments in Sec.~\ref{subsec:exp:instruction}), which our \alg~algorithm can use as input to a powerful LLM to produce the domain of prompts via \emph{in-context learning} \cite{ArXiv23_lin2023use}.
As another example, the initial task description from the user may also be an initial prompt for the task (we follow this in our experiments in Sec.~\ref{subsec:exp:image}), and our \alg~algorithm uses a powerful LLM (e.g., ChatGPT) to rephrase this initial prompt to produce the domain of prompts.
This renders our \alg~algorithm highly flexible and versatile across a broad spectrum of real-world applications.

    \section{Experiments}
    \label{sec:experiments}
\vspace{-2mm}
We test the performance of our \alg~using 3 sets of tasks: optimization of user instructions (Sec.~\ref{subsec:exp:instruction}), prompt optimization for text-to-image generative models (Sec.~\ref{subsec:exp:image}), and response optimization with human feedback (Sec.~\ref{subsec:exp:response:opt}).
To the best of our knowledge, our \alg~is the first algorithm that is designed to efficiently solve the problem of POHF.
We compare our \alg~with 3 natural baseline methods which we adapt to POHF.
(1) \textbf{Random Search} randomly selects a prompt in every iteration and hence ignores the preference feedback.
(2) \textbf{Linear Dueling Bandits} \cite{bengs2022stochastic} uses a linear function to model the latent score function $u$ and adopts a strategy from \cite{bengs2022stochastic} to select the pair of prompts 
(more details in App.~\ref{app:sec:theoretical:justification}). After every iteration, the prompt predicted by the linear model to achieve the largest score is reported as the best prompt.
(3) \textbf{Double Thompson Sampling (DoubleTS)} was recently applied to the problem of response optimization with human feedback by 
\cite{arXiv23_dwaracherla2024efficient} and was shown to be the best-performing method. 
We follow the implementation of DoubleTS from \cite{arXiv23_dwaracherla2024efficient}: We choose the pair of prompts by independently running Thompson sampling (TS) twice, in which the reward/score uncertainty is modeled using Epistemic NNs (which consists of 10 individual MLPs). We also use TS to choose the prompt to report as the best prompt after every iteration.
Note that DoubleTS incurs significantly more computational costs than our \alg, mainly because DoubleTS needs to train $10$ MLPs (in contrast to $1$ MLP needed by our \alg) in every iteration.

\subsection{Optimization of User Instructions}
\label{subsec:exp:instruction}
\vspace{-0.5mm}
To begin with, we simulate real-world scenarios in which a user aims to find the optimal instruction for a task while only giving human preference feedback.
We adopt $30$ instruction induction tasks from~\cite{ArXiv23_chen2023instructzero,ArXiv23_lin2023use}, which have been commonly used by previous works on instruction optimization for black-box LLMs \cite{ArXiv23_chen2023instructzero,hu2024localized,ArXiv23_lin2023use}. 
For every task, a dataset of input-output exemplars is available, which we use to simulate the human preference feedback. Specifically, for selecting every pair of instructions/prompts $x_{t,1}$ and $x_{t,2}$, we use the validation dataset for this task to calculate the validation accuracy achieved by both instructions, which we adopt as their ground-truth latent score values: $u(x_{t,1})$ and $u(x_{t,2})$. 
Then, we calculate the preference probability $\mathbb{P}(x_{1} \succ x_{2}) = \sigma (u(x_{1}) - u(x_{2}))$, and use it as the probability in a Bernoulli distribution to sample the binary preference observation $y_t = \mathbbm{1}(x_{t,1} \succ x_{t,2})$.
This also naturally allows us to report the validation accuracy achieved by an instruction $x$ as its corresponding latent score value $u(x)$, which we plot in our results (Fig.~\ref{fig:benchmark-db}).
Of note, unlike the previous works \cite{ArXiv23_chen2023instructzero,hu2024localized,ArXiv23_lin2023use}, the validation dataset for each task is not used by our algorithm; instead, it is only used to simulate the human preference feedback.

Here, we consider the scenario where the user 
provides a small number of input-output exemplars as the initial task description (Fig.~\ref{fig:problem:setting}), and we use these exemplars to generate the domain of prompts for our \alg~via in-context learning (Sec.~\ref{subsec:method:user:pref:feedback}).
Specifically, to generate each prompt/instruction in the domain, we randomly sample 5 exemplars from the dataset of 100 exemplars (which are separate from the validation set), and ask ChatGPT to generate the instruction that best describes the input-output relationship of these 5 exemplars via in-context learning.
We provide the ChatGPT template used here in Example~\ref{induction-template} (App.~\ref{app:subsec:additional:exp:details}).
Fig.~\ref{fig:benchmark-db} displays the performances of different methods averaged over $30$ tasks.
After each iteration, every method reports a prompt as the best prompt, and its corresponding latent score (i.e., validation accuracy in this case) is plotted in Fig.~\ref{fig:benchmark-db}.
The figure shows that our \alg~algorithm consistently and significantly outperforms the other methods.
We also demonstrate the progression of the best instruction discovered by our \alg~in Table~\ref{table:instruction-optimization-demo}, which further illustrates the capability of our \alg~to efficiently find good instructions using only preference feedback.

\subsection{Prompt Optimization for Text-to-Image Generative Models}
\label{subsec:exp:image}
\vspace{-0.5mm}
Modern text-to-image generative models, such as DALLE-3 \cite{betker2023improving}, have shown remarkable capabilities in generating visually appealing images \cite{chen2024evaluating,rombach2022high,song2020denoising}.
These models take a text prompt as input  and generate a corresponding image.
When a user adopts DALLE-3 to generate an image, they may need to manually try a number of different prompts in order to obtain a desirable image.
Interestingly, in such applications, our \alg~algorithm can also be adopted to efficiently find the best prompt for a user.
Specifically, in every iteration, we can use our \alg~algorithm to select a pair of text prompts and generate two corresponding images using DALLE-3, and then ask the user for preference feedback between the two images.
We simulate such scenarios using the experiments in this section.

To begin with, we adopt an initial prompt that describes a complex scene using several sentences 
(see App.~\ref{table:init-instruction-image-gen} for more details),
and rephrase the initial prompt to produce a large number of text prompts (more details in App.~\ref{appendix:additional-details}).
These prompts are used as the domain of prompts for our \alg, and we select one of the prompts from the domain as the ground-truth prompt. Our implicit assumption is that \emph{the image generated by this ground-truth prompt is the image which is most desirable by the user}. 
Therefore, for every candidate prompt $x$ in the domain, we measure the similarity of its generated image with the image generated by the ground-truth prompt and use the similarity as the latent score $u(x)$ of this prompt.
As a result, for every pair of selected prompts $x_{t,1}$ and $x_{t,2}$, we can calculate their preference probability using the BTL model: $\mathbb{P}(x_{1} \succ x_{2}) = \sigma (u(x_{1}) - u(x_{2}))$, and then sample a binary preference observation $y_t$ from a Bernoulli distribution with probability $\mathbb{P}(x_{1} \succ x_{2})$.
In this case, the goal of our \alg~is to efficiently find a prompt to produce an image that is most preferred by a user, while only requiring a small number of user preference feedback instances.

\begin{figure}[!ht]
    \centering
    \includegraphics[width=0.245\linewidth]{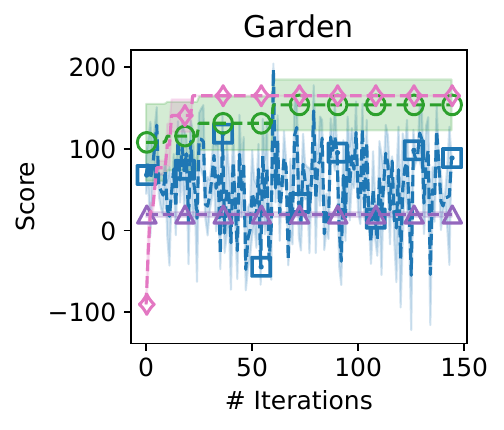}
    \includegraphics[width=0.245\linewidth]{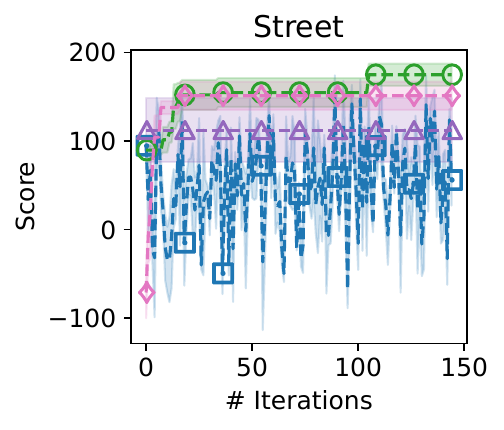}
    \includegraphics[width=0.245\linewidth]{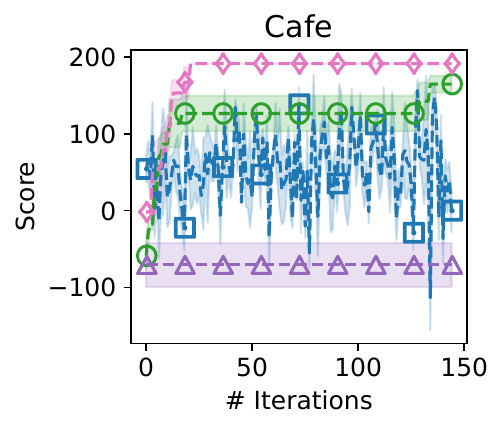}
    \includegraphics[width=0.245\linewidth]{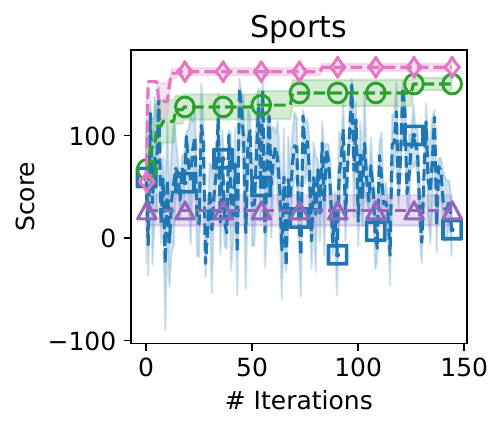}
    \includegraphics[width=0.9\linewidth]{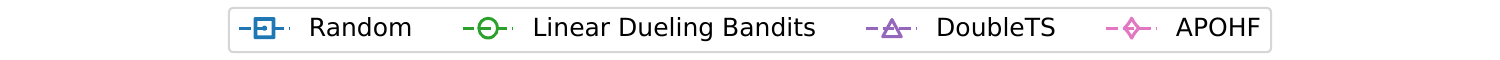}
    \caption{
    Performances in
    prompt optimization for image generation in Sec.~\ref{subsec:exp:image} (4 different scenes).}
    \label{fig:image-gen}
    \vspace{-2mm}
\end{figure}

\begin{figure}
    \centering
    \includegraphics[width=0.9\linewidth]{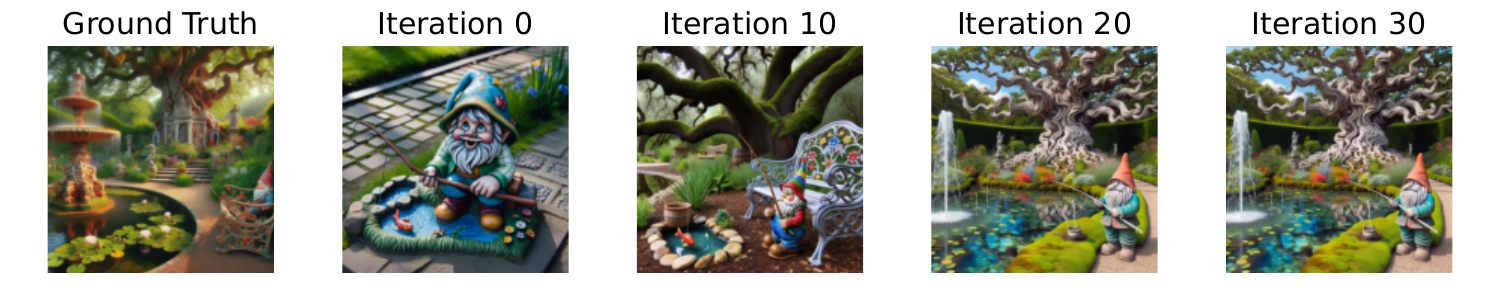}
    \includegraphics[width=0.9\linewidth]{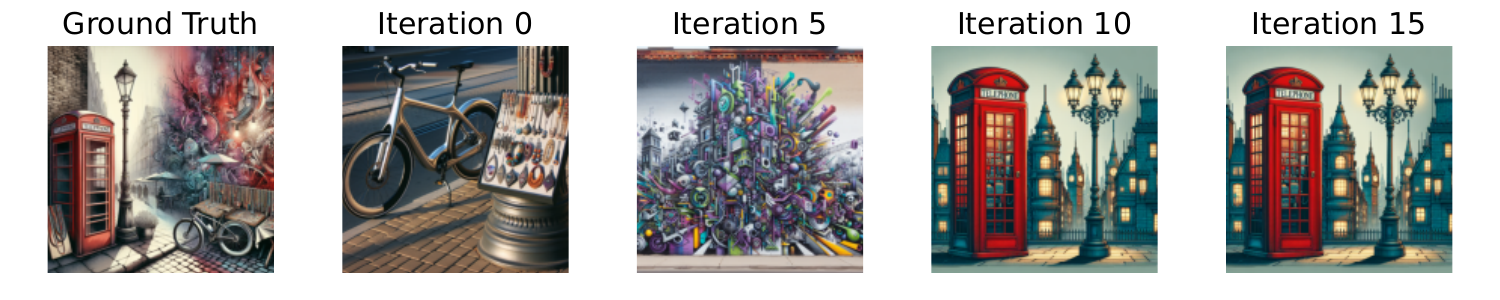}
    \includegraphics[width=0.9\linewidth]{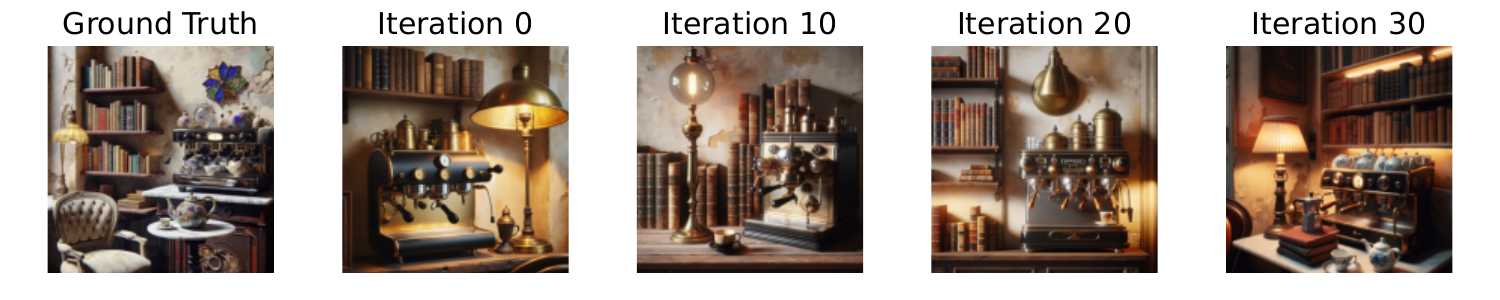}
    \includegraphics[width=0.9\linewidth]{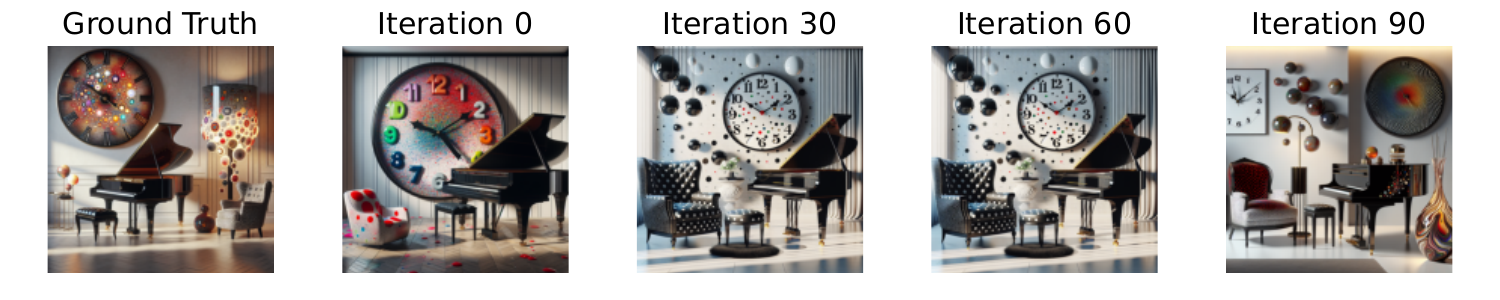}
        \caption{Images generated by the best prompt discovered by our \alg~across different iterations.}
    \label{fig:image-gen:demos}
\end{figure}

We repeat the experiment for 4 different scenes and report the scores of different methods in Fig.~\ref{fig:image-gen}.
The results show that our \alg~consistently outperforms the other baselines across different scenes.
That is, our \alg~is able to efficiently discover a prompt to generate an image that satisfies the user's preferences.
We also demonstrate in Fig.~\ref{fig:image-gen:demos} the evolution of the images generated by the best prompts discovered by our \alg~across different iterations. The results suggest that as more user feedback is collected, \emph{our \alg~can efficiently produce images which better align with the 
image the user has in mind}.
Note that here we intend for the generated images to match the high-level semantic information of the ground-truth image rather than the image details, which are usually uncontrollable due to the inherent randomness in image generation.
This experiment showcases the considerable potential of our \alg~beyond text-generation tasks, suggesting its applicability to a wide range of multi-modal tasks where using human feedback is preferable.

\vspace{-2mm}

\subsection{Response Optimization with Human Feedback}
\label{subsec:exp:response:opt}
In addition to adapting the response of an LLM by optimizing the prompt (i.e., by solving POHF), our \alg~algorithm can also be used to further refine the response from the LLM by tackling the problem of \emph{response optimization with human feedback} (Sec.~\ref{sec:introduction}).
Specifically, given a prompt from a user, we can let the LLM generate a large pool of responses and then try to choose the best response from the pool.
Similar to POHF, instead of requesting the user for a numeric score, it is much easier to ask the user for preference feedback between a pair of responses (Sec.~\ref{sec:introduction}).
This problem setting has also been adopted by the recent work of \cite{arXiv23_dwaracherla2024efficient}.

This problem can be tackled by a \emph{contextual} variant of our \alg. 
That is, every prompt $p$ can be seen as a \emph{context}, and the pool of responses $r$'s generated from this prompt can be considered the domain of \emph{actions}. 
Here, we need to make an important modification to our \alg.
That is, in iteration $t$ after receiving the prompt $p_t$, every input $x$ in the domain is now the embedding of the concatenation of the prompt $p_t$ and one of the LLM-generated responses $r$, which we denote as $x=[p_t,r]$. 
As an implication, the domain $\mathcal{X}_t$ from which we choose a pair of inputs changes in every iteration (as a result of the changing prompt $p_t$).
However, the strategy for selecting the pair of inputs remains the same (Sec.~\ref{subsec:method:selecting:pair:of:prompts}), except that the fixed domain $\mathcal{X}$ 
is now replaced by the changing domain $\mathcal{X}_t$.

\begin{wrapfigure}{R}{0.4\textwidth}
\begin{minipage}{0.4\textwidth}
\vspace{-8mm}
\begin{figure}[H]
    \centering
    \includegraphics[width=0.76\linewidth]{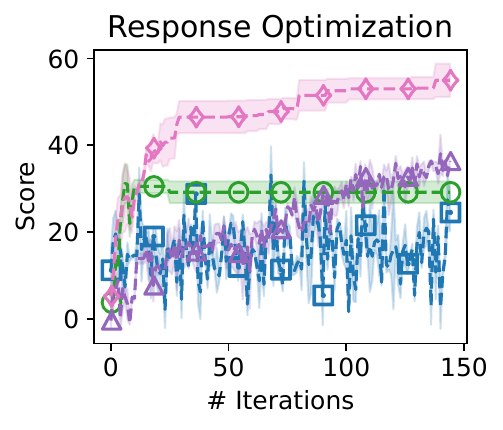}\\
    \includegraphics[width=0.9\linewidth]{figures/legend_figure-2rows.pdf}
    \caption{Scores of different methods for response optimization (Sec.~\ref{subsec:exp:response:opt}).}
    \label{fig:response-selection}
\end{figure}
\vspace{-10mm}
\end{minipage}
\end{wrapfigure}
To simulate the user preferences between different responses, we adopt the same approach as \cite{arXiv23_dwaracherla2024efficient}.
That is, we use a reward model which is pre-trained using the Anthropic Helpfulness and Harmlessness datasets \cite{bai2022training}. 
Then, given a user prompt $p_t$, for every LLM-generated response $r$, we use the output from the pre-trained reward model as the latent score value $u([p_t,r])$ for this prompt-response pair.~Then, for every pair of selected responses $r_{t,1}$ and $r_{t,2}$ by our \alg, we can calculate the preference probability following the BTL model $\Prob{r_{t,1} \succ r_{t,2}} = \sigma (u([p_t,r_{t,1}]) - u([p_t,r_{t,2}]))$ and then use it to sample a binary preference observation $y_t$.
The results are shown in Fig.~\ref{fig:response-selection}, in which our \alg~significantly outperforms the other methods, including DoubleTS, which is found to be the best-performing method in \cite{arXiv23_dwaracherla2024efficient}.
We also show an example of how the response optimized by our \alg~is improved across iterations in Table \ref{table:response-selection-demo-2}.
The response discovered by our \alg~after only 20 iterations is both well organized (via a numbered list) and detailed, which aligns well with human preferences.
This demonstrates the ability of our \alg~to \emph{further refine the response of an LLM to make it more preferable for human users}, while only requiring human preference feedback.

\begin{table}
    \centering
  \begin{minipage}[t]{0.52\textwidth}
    \caption{
    The best instructions selected by our \alg~in different iterations (Sec.~\ref{subsec:exp:instruction}). Full table can be found in~\cref{table:instruction-optimization-demo-full}.
    }
    \label{table:instruction-optimization-demo}
    \begin{center}
    \resizebox{\textwidth}{!}{
    \begin{tabular}{m{1.2cm}|m{0.38cm}|m{7cm}|c}
    \hline
     \textbf{Task} & \textbf{Iter} & \centering \textbf{Instruction} & \textbf{Score} \\
    \hline
    \multirow{3}{=}{antonyms}& 0 &  add the prefix "un-" to the given words to form their opposites. & 0.45 \\
    \cline{2-4}
    & 5 &  remove the "un-" prefix from each word. & 0.45 \\
    \cline{2-4}
    & 10 &  provide the opposite of the given words. & 0.70 \\
    \hline
    \multirow{3}{=}{larger animal}& 0 &  choose the second animal in each pair, so the output is the second animal in each pair. & 0.30 \\
    \cline{2-4}
    & 60 &  choose the animal that is not a type of fish, and if both animals are not fish, then choose the first animal. & 0.55 \\
    \cline{2-4}
    & 120 &  choose the animal that is larger in size, so the output would be:
    wildebeest
    shark
    elk
    pit bull
    manatee & 1.00 \\
    \hline
    \multirow{3}{=}{sentiment}& 0 &  provide negative responses to the given inputs. & 0.00 \\
    \cline{2-4}
    & 60 &  provide an output based on the given input. & 0.00 \\
    \cline{2-4}
    & 120 &  provide the sentiment (positive/negative) of the given inputs. & 0.90 \\
    \hline
    \multirow{3}{=}{word sorting}& 0 & "Please alphabetize the following list of words." & 0.40 \\
    \cline{2-4}
    & 30 &  rearrange the words in the list in alphabetical order and the output provided is the rearranged list of words. & 0.75 \\
    \cline{2-4}
    & 60 &  rearrange the words in the list in alphabetical order and output the sorted list. & 0.85 \\
    \hline
    \end{tabular}}
    \end{center}
  \end{minipage}
  \hfill
  \begin{minipage}[t]{0.46\textwidth}

    \caption{
    The best responses selected by our \alg~across iterations (Sec.~\ref{subsec:exp:response:opt}). The prompt is ``Human: What is there to do in Atlantic City?''.
    We omit some words here (replaced by ``..'') for shorter display. Full table in App.~\ref{table:response-selection-demo-2-full}.
    }
    \label{table:response-selection-demo-2}
    \begin{center}
    \resizebox{\textwidth}{!}{
    \begin{tabular}{c|m{8cm}|c}
    \hline
     \textbf{Iter} &\centering \textbf{Response} & \textbf{Score} \\
    \hline
     0 & There are many activities to do in Atlantic City! You can visit the famous Atlantic City Boardwalk, try your luck at the casinos, enjoy the beach and water activities, shop at the outlets and specialty stores, dine at top-rated restaurants, enjoy live entertainment and nightlife, visit museums and historical sites, and attend events and festivals throughout the year. & 81.96 \\
    \hline
    10 & There are plenty of things to do in Atlantic City! You can visit the famous boardwalk, try your luck at the casinos, enjoy a show or concert at one of the many entertainment venues, dine at a variety of restaurants, shop at the outlets and specialty stores, relax on the beach, or take a stroll through the historic district. There's something for everyone in Atlantic City! & 91.46\\
    \hline
    20 & There are many attractions and activities in Atlantic City, including:
    
    1. Casinos: Atlantic City is known for its many casinos, where you can try your luck at slots, poker, blackjack, and more.
    2. Boardwalk: ...
    3. Beach: ..
    4. Steel Pier: ...
    5. Shows and concerts: ...
    6. Nightlife: ...
    7. Dining: ...
    8. Shopping: ... & 180.14
    \\
    \hline
    \end{tabular}}
    \end{center}
    
  \end{minipage}
\end{table}

\vspace{-2mm}
\section{Ablation Study}
\vspace{-2mm}
Here we perform ablation study of our \alg~using the experiments in Sec.~\ref{subsec:exp:response:opt}.

\begin{figure}
\vspace{-4mm}
    \centering
  \begin{minipage}[b]{0.49\textwidth}
    \centering
    \includegraphics[width=0.65\linewidth]{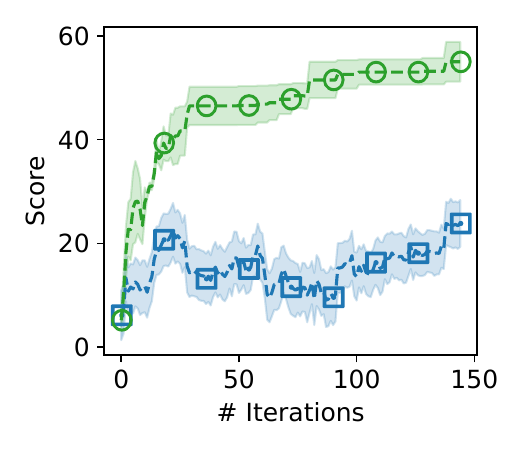}
    \\
    \vspace{-2mm}
    \includegraphics[width=0.65\linewidth]{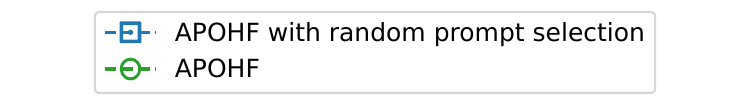}
    \vspace{-1.5mm}
    \caption{Comparison of 
    our arm selection strategy with random selection.}
    \label{fig:ablation:arm:selection:strategy}
  \end{minipage}
  \hfill
  \begin{minipage}[b]{0.49\textwidth}
    \centering
    \includegraphics[width=0.63\linewidth]{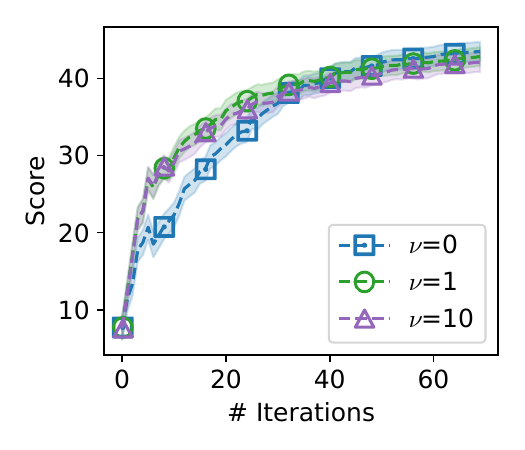}
    \caption{Comparison of the performance of our \alg~algorithm with different values of $\nu$ (i.e., the exploration parameter).}
    \label{fig:ablation:exploration}
  \end{minipage}
  \vspace{-6mm}
\end{figure}

\textbf{Effectiveness of Our Prompt Selection Strategy.}
Here, we further verify the effectiveness of our theoretically principled prompt selection strategy.
We replace the strategy of our \alg~to select a pair of prompts by uniform random selection while keeping all other components of our \alg~fixed.
That is, after every iteration, we still train the NN using the current history of observations
as described in Sec.~\ref{subsec:method:train:nn}, and report the prompt maximizing the prediction of the NN as the best prompt.
The results (Fig.~\ref{fig:ablation:arm:selection:strategy}) show that randomly selecting the pair of prompts significantly degrades the performance of our \alg, further validating the effectiveness of our prompt selection strategy (Sec.~\ref{subsec:method:selecting:pair:of:prompts}).

\textbf{Impact of the Exploration Parameter.}
Here, we examine the impact of the exploration parameter $\nu$ on our \alg~algorithm.
The results (Fig.~\ref{fig:ablation:exploration}) show that setting $\nu=0$ (i.e., not performing any exploration) degrades the performance of our \alg.
This is because it limits the ability of our \alg~to sufficiently explore the space of possible prompts.
On the other hand, using a large value of $\nu=10$ does not significantly affect the performance of \alg.
This is because although a large $\nu$ may result in excessive exploration when selecting the second prompt, the value of $\nu$ does not alter our strategy to choose the first prompt. Therefore, a large exploration parameter $\nu$ does not significantly diminish the ability of our \alg~to exploit the prediction of the NN.

\begin{figure}[!ht]
    \centering
    \includegraphics[width=0.99\linewidth]{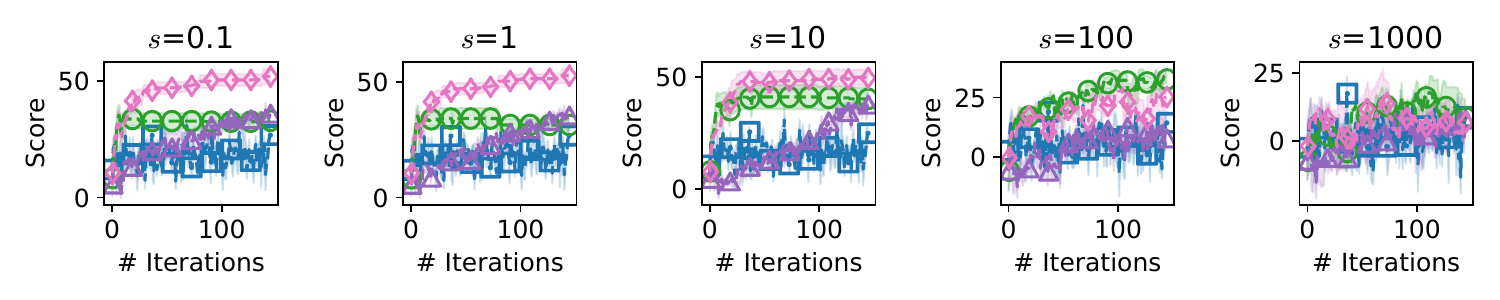}
    \\
    \includegraphics[width=0.9\linewidth]{figures/legend_figure.pdf}
    \caption{Comparison of the performances of different algorithm under different levels of noise in human feedback. 
    Here $s$ controls the level of noise, such that a larger $s$ results in a higher noise level.}
    \label{fig:ablation:noise}
    \vspace{-3mm}
\end{figure}

\textbf{Impact of the Level of Noise in Preference Feedback.}
Here, we study the impact of the level of noise in preference feedback on the performance of different algorithms.
We alter the level of noise in preference feedback by adjusting the scale of the latent score function $u$. 
A smaller scale of the scores results in noisier preference observations and hence leads to a more difficult optimization problem.
This is because according to the BTL model $\mathbb{P}(x_{1} \succ x_{2}) = \sigma (u(x_{1}) - u(x_{2}))$, a smaller scale of $u(\cdot)$ generally makes the preference probability closer to $0.5$.
This renders the resulting binary observation $y_t = \mathbbm{1}(x_{t,1} \succ x_{t,2})$ more similar to a purely random sample (with a probability of $0.5$) and hence noisier.
The results (Fig.~\ref{fig:ablation:noise}) verify that the smaller the noise, the more pronounced the advantage of our \alg. Meanwhile, as the noise level becomes too large, the problem becomes excessively difficult for all methods, and eventually, all algorithms achieve similar performances.

    \section{Related Work}
    \label{sec:related_work}

\vspace{-2mm}
Prompt optimization, also referred to as \emph{instruction optimization}, has been gaining popularity thanks to its ability to improve the performance of LLMs without 
parameter fine-tuning.
Earlier works aimed to optimize the prompt for white-box LLMs, such as AutoPrompt \cite{EMNLP20_shin2020autoprompt}, FluentPrompt \cite{Shi2022TowardHR}, as well as other works based on soft prompt \cite{EMNLP21_lester2021power,ACL21_li2021prefix,zhong2021optiprompt}.
Recently, more focus has been shifted to optimizing the prompt for black-box LLMs.
Among them, BBT \cite{sun2022black}, BBTv2 \cite{sun2022bbtv2} and Clip-Tuning \cite{chai2022clip} require access to the input embedding and output logits of the black-box LLM.
Other recent works have removed this restriction.
For example, GRIPS \cite{ACL23_prasad2023grips} and APO \cite{ArXiv23_pryzant2023automatic} used edit-based operations to select candidate prompts for prompt optimization.
Other works have adopted evolutionary algorithms (e.g., EvoPrompt \cite{guo2023connecting} and Promptbreeder \cite{fernando2023promptbreeder}), reinforcement learning (e.g., BDPL \cite{diao2023blackbox} and PRewrite \cite{kong2024prewrite}), and planning-based methods (e.g., PromptAgent \cite{wang2023promptagent}) to achieve prompt optimization for black-box LLMs.
The work of \cite{ICLR23_zhou2022large} proposed APE, which generates candidate instructions using an LLM and selects those high-scoring candidates for further refinement.
The OPRO algorithm \cite{yang2023large} was developed to use an LLM to solve generic black-box optimization problems and was applied to the problem of prompt optimization. 
The work of \cite{manas2024improving} introduced OPT2I, which uses an LLM to sequentially revise the prompt for text-to-image generative models, in order to maximize a score measuring the consistency of the generated image with the given prompt.

Some recent works have tackled prompt optimization for black-box LLMs by converting it to a continuous optimization problem. 
InstructZero \cite{ArXiv23_chen2023instructzero} adopted a separate white-box LLM to convert prompt optimization to optimizing the soft prompt and used Bayesian optimization to solve the resulting continuous optimization problem.
INSTINCT \cite{ArXiv23_lin2023use} used neural bandits to sequentially select the instructions to query and leveraged the strong expressive power of neural networks to achieve better function modeling and hence better prompt optimization.
ZOPO \cite{hu2024localized} adopted zeroth-order optimization (ZOO) while estimating the gradient based on a neural network, and further improved the performances of InstructZero and INSTINCT.
In addition, \cite{shi2024best} demonstrated the potential of drawing inspirations from best arm identification for prompt optimization, and \cite{chen2024online} used neural bandits for personalized content generation using white-box LLMs.
Importantly, to the best of our knowledge, \emph{these previous works are not able to tackle the problem of POHF considered in our work}, because they require a numeric score to evaluate the efficacy of each prompt.

RLHF has become the most widely used method for aligning the responses of LLMs with human values \cite{dubois2024alpacafarm,ouyang2022training,ziegler2019fine}. More comprehensive discussions on RLHF can be found in recent surveys \cite{casper2023open,chaudhari2024rlhf}.
More recently, some methods have been developed to sidestep the need for RL and directly use a preference dataset for alignment, including direct preference optimization (DPO) \cite{rafailov2024direct}, SLiC \cite{zhao2023slic}, as well as other extensions \cite{amini2024direct,azar2024general,gou2024mixed,liu2024lipo,morimura2024filtered,tang2024generalized,wang2023beyond}.
The recent work of \cite{arXiv23_dwaracherla2024efficient} has shown the potential of efficient exploration methods to improve the response of LLMs with human preference feedback.

    \section{Conclusion and Limitations}
    \label{sec:conclusion}
\vspace{-1.5mm}

We have introduced the problem of POHF, in which our goal is to optimize the prompt for black-box LLMs while using only human preference feedback.
To address POHF, we have proposed the \alg~algorithm, which uses a neural network trained using preference feedback to model the latent score function, and chooses the pair of prompts to query based on a principled strategy inspired by dueling bandits.
By using various tasks, including user instruction optimization, prompt optimization for text-to-image generative models, and response optimization with human feedback, we empirically validate that our \alg~is able to find a good prompt for a task using a small number of human feedback instances.
A potential limitation of our \alg~is that it currently does not accommodate the scenario where more than 2 prompts are selected in every iteration, and the user provides feedback regarding the ranking of the responses from these prompts. We plan to tackle this in future work by developing novel and theoretically principled strategies to choose more than 2 prompts to query.

    \bibliographystyle{abbrv} 
    \bibliography{ref}

    \newpage
    \appendix

\section{Addtional Details for Experiments}\label{appendix:additional-details}
\subsection{License for datasets}
(1) Instruction induction dataset~\cite{ArXiv23_chen2023instructzero,ArXiv23_lin2023use} for optimizing the user instruction: MIT License; (2) Anthropic Helpfulness and Harmlessness datasets~\cite{bai2022training} for response optimization: MIT License.

\subsection{Computational resources}
All the experiments are run on a server with AMD EPYC 7763 64-Core Processor, $1008$GB RAM, and $8$ NVIDIA L40 GPUs.
\subsection{Additional details on experimental settings}\label{app:subsec:additional:exp:details}

\paragraph{Hyper-parameters.} 
We use an MLP with $2$ hidden layers as the NN for the latent score prediction. Each hidden layer has a width of $32$. At each iteration of our \alg~we re-initialize the NN and train the NN using all available human feedback data for $1000$ epochs with Adam optimizer and a learning rate of $0.001$. We run all algorithms for $150$ iterations. We normalize the score distributions for all applications to $\mathcal{N}(0, 100)$ such that the simulated feedback obtained by the BTL model will not be too noisy. We use the hyper-parameters of $\nu=1$ and $\lambda=0.1$ for our \alg~and Linear Dueling Bandits. For the prompt optimization for text-to-image generative models, we use a larger $\nu=10$ for both algorithms for better exploration. All the experiments are run at least $2$ times to obtain the error bars and the average performances. For ChatGPT queries used in all experiments, we use the specific version of ``gpt-3.5-turbo-1106'' API provided by OpenAI.
\paragraph{User instruction optimization.}
We generate a prompt domain with $200$ prompts/instructions. The validation dataset has a size of $20$. The exemplar dataset provided by the user has a size of $100$. The validation accuracy for a prompt/instruction is evaluated by using the validation dataset and querying ChatGPT, which is the same as previous works~\cite{ArXiv23_chen2023instructzero,ArXiv23_lin2023use}. We use MPNet~\cite{song2020mpnet} to obtain the representations of the prompts to be the inputs to our NN for the latent score prediction. 
\paragraph{Prompt optimization for text-to-image generative models.} We generate a prompt domain with $200$ prompts. Specifically, we use the template in Example~\ref{img-induction-rephrase-1} to rephrase the initial prompt for each scene in~\cref{table:init-instruction-image-gen} to obtain the ground-truth prompt. We use the template in Example~\ref{img-induction-rephrase-1} to rephrase the initial prompt again to obtain $10$ different prompts as good candidates in the prompt domain. This is to make sure that the domain contains some prompts that are very close to the ground-truth prompt. For the generation of the other $190$ prompts in the domain, we first select a subset of sentences from the initial prompt. Specifically, each sentence in the initial prompt is selected with a probability of $0.3$ independently. This is to simulate real-life scenarios where the prompts provided by the users may only contain a fraction of the information needed to generate the ground-truth or ideal images. We combine the selected subset of sentences to form a new prompt and use the template in Example~\ref{img-induction-rephrase-2} to rephrase it to obtain a new element in the prompt domain. We repeat the above procedures to obtain the other $190$ prompts. We use the  DALLE-3  model with the generation quality as ``standard'' and the generation size as ``$1024\times1024$''. We use CLIP~\cite{radford2021learning} to obtain the representations of the ground-truth image and the generated images. We use the cosine similarity function to calculate the similarity score between the representations of the ground-truth image and the generated image as the quality measure for the corresponding generated image. We use vision transformer~\cite{dosovitskiy2020image} to obtain the representations of the generated images to be the inputs to the NN for the latent score prediction. The reason for using a different representation model for the latent score prediction is to simulate real-life scenarios in which we do not have prior knowledge about the ground-truth score function. 
\paragraph{Response optimization with human feedback.} We randomly select $10$ questions from the test dataset of the Anthropic Helpfulness dataset as the prompts. For each prompt, we generate $50$ responses from ChatGPT. We set the temperature parameter of ChatGPT to be $1.0$ so that the generated responses are different from each other. We use a fine-tuned GPT-2 model~\cite{radford2019language} to obtain the ground-truth scores for prompt-response pairs. Specifically, the GPT-2 is fine-tuned on Anthropic Helpfulness dataset~\cite{bai2022training} to determine the helpfulness of the response w.r.t. a prompt by outputting a score. For each response, we concatenate its corresponding prompt as the prefix and input it to the fine-tuned GPT-2 model to obtain the ground-truth score. We use MPNet to obtain the representations of the prompt-response pairs to be the inputs to our NN for the latent score prediction. For each iteration of the algorithm, a prompt is selected in a round-robin fashion with a fixed order. This is for the purpose of result visualization and fair comparison since different algorithms select responses for the same prompt in each corresponding iteration, and their performances are evaluated based on the same domain in this way. The score for each algorithm in an iteration in~\cref{fig:response-selection} is calculated by using the trained latent score prediction model in this iteration from each algorithm to select the best responses for each of the $10$ prompts and evaluating these $10$ prompt-response pairs with the ground-truth score function to obtain an average score.

\begin{mycolorbox}[label=induction-template]{Query}{Instruction Induction Template}    
    Input: [INPUT] \\
    Output: [OUTPUT] \\\\
    <More exemplars...> \\\\
    Input: [INPUT] \\
    Output: [OUTPUT] \\\\
    The instruction was to: \\

\end{mycolorbox}
\begin{table}[t]
\caption{
Initial 
prompts
for generating images with different scenes.
}
\label{table:init-instruction-image-gen}
\begin{center}
\begin{tabular}{l|m{12cm}}
\hline
 \textbf{Scene} & \textbf{Prompt} \\
\hline
Garden & In a vibrant garden, a grand marble fountain gushes clear water, dazzling in the sunlight. Nearby, a centuries-old oak tree stands with sprawling, gnarled branches. A vintage wrought iron bench with floral patterns offers a quaint seat. Beside the path, a whimsical, brightly painted gnome statue holds a fishing rod towards a small pond. In the pond, lily pads float with blooming white lilies. \\
\hline
Street & On a lively city street, a striking vintage red telephone booth pops against the muted city colors. Nearby, a vibrant graffiti mural adds color to a plain brick wall, featuring an abstract mix of urban elements. A futuristic bicycle with a shiny, aerodynamic silver frame is locked to a lamppost. A small vendor's stall on the sidewalk displays handmade, colorful beaded jewelry, glistening in the afternoon sun. In the background, an ornate old-fashioned street lamp emits a warm glow as dusk approaches. \\
\hline
Cafe & In a quaint cafe corner, a vintage espresso machine with polished brass fixtures and a matte black body gleams under an antique lamp. A rustic wooden bookshelf, brimming with well-worn books, stands against a distressed cream wall. A marble table at the room's center holds a delicate porcelain teapot with intricate blue flowers, from which steam gently rises. Beside the table, a colorful glass mosaic cat sculpture perches on a mismatched velvet chair, casting playful reflections around. \\
\hline
Sports & A sleek grand piano with a glossy black surface speckled with white spots stands at the room's center. On the wall, a colorful clock features a face marked by vibrant, multicolored spots for each hour. Beside it, a tall floor lamp sports a leopard-spot patterned lampshade in black and gold. A plush armchair in the corner showcases bold red polka dots on a white background. On a nearby table, a delicate glass vase captivates with swirling, iridescent spots that shimmer in the light. \\
\hline
\end{tabular}
\end{center}
\end{table}

\begin{mycolorbox}[label=img-induction-rephrase-1]{Query}{Image Generation Instruction Rephrasing Template 1}    
    Rephrase the following description: [Initial instruction] \\
    The rephrased description is: 
\end{mycolorbox}

\begin{mycolorbox}[label=img-induction-rephrase-2]{Query}{Image Generation Instruction Rephrasing Template 2}    
    Modify the following description by rephrasing and changing some information: [Initial instruction] \\
    The modified description is: 
\end{mycolorbox}

\section{Addtional Experimental Results}\label{appendix:additional-results}

\cref{table:response-selection-demo-1-full} and~\cref{table:response-selection-demo-2-full} show the best responses selected by our \alg~in different iterations. An observation is that both the response quality and the score increase when more iterations are performed. Specifically, for~\cref{table:response-selection-demo-1-full}, the response in iterations $10$ and $20$ include more information on the affordability of different kinds of insurance compared to iteration $0$. For~\cref{table:response-selection-demo-2-full}, as more iterations of our \alg~are performed, the response provides more details on the things to do in Atlantic City. In iteration $20$, the response even includes bullet points for the recommended activities for the users and hence is the most useful response.

\cref{table:instruction-optimization-demo-full} shows the best instructions selected by our \alg~for different tasks. As more iterations are performed, the instructions become more relevant to the tasks and hence have higher scores. 

\begin{table}[t]
\caption{
The best responses selected by our \alg~in different iterations. The prompt is ``Human: Should you buy life insurance?
Assistant: I’m sure life insurance is useful for a lot of people. I can also tell you how you can get insurance, depending on your income and needs.
Human: Is term or whole life insurance better?''
}
\label{table:response-selection-demo-1-full}
\begin{center}
\resizebox{\textwidth}{!}{
\begin{tabular}{c|m{12cm}|c}
\hline
 \textbf{Iter} & \centering \textbf{Response} & \textbf{Score} \\
\hline
0 & It really depends on your individual circumstances and financial goals. Term life insurance provides coverage for a specific period of time, while whole life insurance provides coverage for your entire life and also has an investment component. It's important to carefully consider your financial situation and consult with a financial advisor before making a decision. & 45.54 \\
\hline
10 & Term life insurance is typically more affordable and provides coverage for a specific period of time, while whole life insurance provides coverage for your entire life and also includes a cash value component. The best option for you will depend on your individual needs and financial situation. It's important to carefully consider your options and consult with a financial advisor or insurance expert. & 100.96\\
\hline
20 & Term life insurance is typically more affordable and provides coverage for a specific period of time, while whole life insurance provides coverage for your entire life and also includes a cash value component. The best option for you will depend on your individual needs and financial situation. It's important to carefully consider your options and consult with a financial advisor or insurance expert. & 100.96\\
\hline
\end{tabular}}
\end{center}
\end{table}

\begin{table}[t]
\caption{
The best responses selected by our \alg~in different iterations. The prompt is ``Human: What is there to do in Atlantic City?''.
}
\label{table:response-selection-demo-2-full}
\begin{center}
\resizebox{\textwidth}{!}{
\begin{tabular}{c|m{12cm}|c}
\hline
\textbf{Iter} & \centering \textbf{Response} & \textbf{Score} \\
\hline
0 & There are many activities to do in Atlantic City! You can visit the famous Atlantic City Boardwalk, try your luck at the casinos, enjoy the beach and water activities, shop at the outlets and specialty stores, dine at top-rated restaurants, enjoy live entertainment and nightlife, visit museums and historical sites, and attend events and festivals throughout the year. & 81.96 \\
\hline
10 & There are plenty of things to do in Atlantic City! You can visit the famous boardwalk, try your luck at the casinos, enjoy a show or concert at one of the many entertainment venues, dine at a variety of restaurants, shop at the outlets and specialty stores, relax on the beach, or take a stroll through the historic district. There's something for everyone in Atlantic City! & 91.46\\
\hline
20 & There are many attractions and activities in Atlantic City, including:

1. Casinos: Atlantic City is known for its many casinos, where you can try your luck at slots, poker, blackjack, and more.

2. Boardwalk: Take a stroll on the iconic Atlantic City Boardwalk, lined with shops, restaurants, and amusement attractions.

3. Beach: Enjoy a day of sun and sand at the Atlantic City beach, a popular spot for swimming, sunbathing, and water sports.

4. Steel Pier: Visit this historic amusement park featuring rides, games, and entertainment for the whole family.

5. Shows and concerts: Catch a live performance or concert at one of the many entertainment venues in Atlantic City, including the Hard Rock Hotel \& Casino and the Borgata Hotel Casino \& Spa.

6. Nightlife: Experience the vibrant nightlife of Atlantic City, with numerous bars, nightclubs, and lounges offering live music, DJs, and dancing.

7. Dining: Indulge in a variety of dining options, from upscale restaurants to casual eateries serving fresh seafood, steaks, and international cuisine.

8. Shopping: Explore the shops and boutiques in Atlantic City, offering everything from designer fashions to unique souvenirs. & 180.14
\\
\hline
\end{tabular}}
\end{center}
\end{table}

\begin{table}[t]
\caption{
The best instructions selected by our \alg~in different iterations.
}
\label{table:instruction-optimization-demo-full}
\begin{center}
\resizebox{\textwidth}{!}{
\begin{tabular}{m{1.5cm}|c|m{9cm}|c}
\hline
 \textbf{Task} & \textbf{Iter} & \centering \textbf{Instruction} & \textbf{Score} \\
\hline
\multirow{3}{=}{antonyms}& 0 &  add the prefix "un-" to the given words to form their opposites. & 0.45 \\
\cline{2-4}
& 5 &  remove the "un-" prefix from each word. & 0.45 \\
\cline{2-4}
& 10 &  provide the opposite of the given words. & 0.70 \\
\hline
\multirow{3}{=}{informal to formal}& 0 &  rephrase the given sentences, so I have provided the rephrased versions of the input sentences as output. If this is not what you were looking for, please provide more specific instructions. & 0.39 \\
\cline{2-4}
& 5 &  rephrase the given sentences using formal language. & 0.44 \\
\cline{2-4}
& 10 &  rephrase each input sentence using a more formal or polite tone. & 0.47 \\
\hline
\multirow{3}{=}{larger animal}& 0 &  choose the second animal in each pair, so the output is the second animal in each pair. & 0.30 \\
\cline{2-4}
& 60 &  choose the animal that is not a type of fish, and if both animals are not fish, then choose the first animal. & 0.55 \\
\cline{2-4}
& 120 &  choose the animal that is larger in size, so the output would be:

wildebeest
shark
elk
pit bull
manatee & 1.00 \\
\hline
\multirow{3}{=}{orthography starts with}& 0 &  identify the word in the sentence that is in Russian, and for the first three sentences, the word "Russian" was correctly identified. However, for the last two sentences, there were no words in Russian, so the output should have been "N/A" or "none." & 0.00 \\
\cline{2-4}
& 20 &  identify the adjective in each sentence. & 0.15 \\
\cline{2-4}
& 40 &  provide the word that starts with the given letter. & 0.80 \\
\hline
\multirow{3}{=}{rhymes}& 0 &  change the first letter of the word to "inv" and then add the rest of the word. & 0.00 \\
\cline{2-4}
& 4 &  find a word that is an anagram of the given word. & 0.00 \\
\cline{2-4}
& 8 &  change the word to a new word that rhymes with the original word. & 0.40 \\
\hline
\multirow{3}{=}{second word letter}& 0 &  "Provide the index of the first occurrence of the letter 'a' in each word." & 0.00 \\
\cline{2-4}
& 2 &  "Provide the index of the first occurrence of the letter 'a' in each word." & 0.00 \\
\cline{2-4}
& 4 &  "Output the second letter of each word." & 1.00 \\
\hline
\multirow{3}{=}{sentiment}& 0 &  provide negative responses to the given inputs. & 0.00 \\
\cline{2-4}
& 60 &  provide an output based on the given input. & 0.00 \\
\cline{2-4}
& 120 &  provide the sentiment (positive/negative) of the given inputs. & 0.90 \\
\hline
\multirow{3}{=}{taxonomy animal}& 0 &  rearrange the words in alphabetical order, so the output for each input would be the words listed in alphabetical order. & 0.00 \\
\cline{2-4}
& 30 &  rearrange the words in alphabetical order, so the output lists the words in alphabetical order. & 0.00 \\
\cline{2-4}
& 60 &  "Output the animals from the given list." & 0.95 \\
\hline
\multirow{3}{=}{word sorting}& 0 & "Please alphabetize the following list of words." & 0.40 \\
\cline{2-4}
& 30 &  rearrange the words in the list in alphabetical order and the output provided is the rearranged list of words. & 0.75 \\
\cline{2-4}
& 60 &  rearrange the words in the list in alphabetical order and output the sorted list. & 0.85 \\
\hline
\end{tabular}}
\end{center}
\end{table}

\section{Theoretical Justifications for Our \alg~Algorithm}
\label{app:sec:theoretical:justification}
Here, we provide theoretical justifications for the prompt selection strategy of our \alg~algorithm, to show that our \alg~algorithm is theoretically principled.
Our goal here is to provide high-level intuitions rather than giving a complete theoretical analysis.

Recall that as we have introduced in Sec.~\ref{subsec:method:selecting:pair:of:prompts}, our \alg~algorithm selects the first prompt greedily by maximizing the predicted score (from the trained NN), and chooses the second prompt by maximizing upper confidence bound, which is a weighted combination of the score prediction and an uncertainty term \eqref{eq:choose:arm:2}.
This strategy is inspired by previous works on linear dueling bandits \cite{bengs2022stochastic,saha2021optimal}.

Here, we adopt the simplifying assumption that the utility/score function $u$ is a linear function: $u(x) = \theta^{\top} x,\forall \mathcal{X}$ with unknown parameter $\theta$.
With this assumption, our prompt selection strategy can be seen as a modified version of the algorithm from \cite{bengs2022stochastic}. 
Therefore, we follow the notations from \cite{bengs2022stochastic} and present below the most important modifications to the theoretical analysis of \cite{bengs2022stochastic}.
We use $z_{1,2}$ to denote the difference between (the features vectors of) a pair of prompts: $z_{1,2} = x_1 - x_2$ and use $z_{t,1,2}=x_{t,1} - x_{t,2}$ to denote the difference between the pair of selected prompts in iteration $t$. The matrix $M_t = \sum^{t}_{s=1} z_{t,1,2}^{\top} z_{t,1,2}$ intuitively characterizes the information collected up to iteration $t$.

With these notations, $\hat{\theta}^{\top} z$ represents the \emph{predicted reward difference} between a pair of prompts $x_1$ and $x_2$, which in our case, corresponds to $h(x_1;\theta_t) - h(x_2;\theta_t)$. Then, $\theta^{\top} z$ represents the ground-truth reward difference.
Following the standard practice of the analysis of bandit algorithms \cite{bengs2022stochastic}, we assume that the validity of the confidence bound providing a theoretical guarantee on the quality of reward difference estimation: $|\theta^{\top} z - \hat{\theta}^{\top} z | \leq \nu \norm{z}_{M_t^{-1}}$.
With these, the \emph{regret} incurred in 
iteration $t$ can be analyzed as:
\begin{equation}
\begin{split}
2 r_t &= u(x^*) - u(x_{t,1}) + u(x^*) - u(x_{t,2})\\
&\stackrel{(a)}{=} \theta^{\top} (x^* - x_{t,1}) + \theta^{\top} (x^* - x_{t,2})\\
&\stackrel{(b)}{=} \theta^{\top} z^*_{t,1} + \theta^{\top} z^*_{t,2}\\
&= (\theta - \hat{\theta}_t)^{\top} z^*_{t,1} + \hat{\theta}_t^{\top} z^*_{t,1} + (\theta - \hat{\theta}_t)^{\top} z^*_{t,2} + \hat{\theta}_t^{\top} z^*_{t,2}\\
&\stackrel{(c)}{\leq} \hat{\theta}_t^{\top} z^*_{t,1} + \nu\norm{z^*_{t,1}}_{M_t^{-1}} + \hat{\theta}_t^{\top} z^*_{t,2} + \nu \norm{z^*_{t,2}}_{M_t^{-1}}\\
&\stackrel{(d)}{\leq} 2 \hat{\theta}_t^{\top} (x^* - x_{t,1}) + 2 \nu \norm{x^* - x_{t,1}}_{M_t^{-1}} + \hat{\theta}_t^{\top} z_{t,1,2} + \nu\norm{z_{t,1,2}}_{M_t^{-1}}\\
&\stackrel{(e)}{\leq} 2 \hat{\theta}_t^{\top} (x_{t,2} - x_{t,1}) + 2 \nu\norm{x_{t,2} - x_{t,1}}_{M_t^{-1}} + \hat{\theta}_t^{\top} (x_{t,1} - x_{t,2}) + \nu\norm{x_{t,1} - x_{t,2}}_{M_t^{-1}}\\
&\leq \hat{\theta}_t^{\top} (x_{t,2} - x_{t,1}) + 3 \nu\norm{z_{t,1,2}}_{M_t^{-1}} \\
&\stackrel{(f)}{\leq} 3 \nu\norm{z_{t,1,2}}_{M_t^{-1}}.
\end{split}
\end{equation}
Step $(a)$ follows because here we have assumed that the score function $u$ is a linear function; in step $(b)$, we have defined $z^*_{t,1}=x^* - x_{t,1}$ and $z^*_{t,2} = x^* - x_{t,2}$; step $(c)$ follows because we have assumed the validity of the confidence bound as described above; 
step $(d)$ follows simply because $z^*_{t,2} = x^* - x_{t,2} = x^* - x_{t,1} + x_{t,1} - x_{t,2} = z^*_{t,1} + z_{t,1,2}$ (we have also made use of the triangle inequality).

\paragraph{Selection of the Second Prompt.}
Step $(e)$ follows from the way the second prompt is selected: $x_{t,2} = {\arg\max}_{x\in\mathcal{X}}\hat{\theta}_t^{\top} x + \nu \norm{x - x_{t,1}}_{M_t^{-1}}$.
This, importantly, is analogous to the way in which our \alg~algorithm selects the second prompt using \cref{eq:choose:arm:2}.
Note that we have replaced the linear score prediction $\hat{\theta}_t^{\top} x$ by the prediction from our NN: $h(x;\theta_t)$.
We have also used the
gradient $\nabla h(x;\theta_t)$ to replace the original feature vector $x$, which is justified by the theory of the neural tangent kernel (NTK), which has shown that $\nabla h(x;\theta_t)$ 
can be used to approximate
the random Fourier features for the NTK \cite{jacot2018neural}.
Also note that compared to the theory of NTK, we have designed our \alg~algorithm to be more practical following the common practice of neural bandits \cite{zhang2020neural,zhou2020neural}. 
Specifically, in the loss function to train our NN \eqref{eq:customized:loss:function}, for the regularization parameter, we have replaced the theoretical choice of $\frac{1}{2} m \lambda \norm{\theta - \theta_0}^2_{2} $ ($m$ is the width of the NN) by simply $\lambda \norm{\theta}^2_{2}$; regarding the random features of the NTK, we have replaced the theoretical choice of $\frac{1}{\sqrt{m}}\nabla h(x;\theta_t)$ by simply $\nabla h(x;\theta_t)$.

\paragraph{Selection of the First Prompt.}
Step $(f)$ results from the way in which the first prompt is chosen: $x_{t,1}={\arg\max}_{x\in\mathcal{X}}\hat{\theta}_t^{\top}x$.
This is analogous to the way in which our \alg~algorithm selects the first prompt: $x_{t,1} = \arg\max_{x\in\mathcal{X}} h(x;\theta_t)$.

The subsequent analysis follows from standard analysis techniques for linear dueling bandits \cite{bengs2022stochastic}.
Therefore, our strategy to select the two prompts is theoretically principled.

Note that in this section, we have provided some high-level theoretical justifications for the prompt selection strategy of our \alg~algorithm. Our prompt selection strategy can, in fact, be seen as a variant of neural dueling bandit algorithms. 

\section{Broader Impacts}\label{appendix:impacts}
We expect our work to have important positive societal impacts. Specifically, our algorithm can be used to automatically optimize the prompt for LLMs while requiring only preference feedback from the human user. So, our algorithm is likely to make LLMs easier to use for users and hence contribute to the easier and wider adoption of LLMs as well as other advanced AI algorithms. This is expected to positively impact society by improving productivity at both the individual and the societal levels.
On the other hand, a potential negative societal impact is that our algorithm may be adopted by malicious users.
These users could intentionally provide misleading preference feedback to the LLM, in order to find inappropriate prompts for tasks associated with malicious intents. For example, malicious attackers could use our algorithm to find prompts for jailbreaking LLMs.
Developing effective safeguarding methods to prevent such potential malicious use presents interesting future research topics.\\

\end{document}